\pdfoutput=1

\documentclass[11pt]{article}

\usepackage[]{acl}

\usepackage{times}
\usepackage{latexsym}

\usepackage[T1]{fontenc}

\usepackage[utf8]{inputenc}

\usepackage{microtype}

\usepackage{inconsolata}

\usepackage{times}
\usepackage{latexsym}
\usepackage{bm}
\usepackage[T1]{fontenc}
\usepackage[utf8]{inputenc}
\usepackage{microtype}
\usepackage{inconsolata}
\usepackage{multirow}
\usepackage{graphicx}
\usepackage{booktabs}  
\usepackage{subcaption} 

\usepackage{xcolor}
\usepackage{hyperref}
\definecolor{green}{HTML}{38761D}
\definecolor{orange}{HTML}{FF8900}
\definecolor{blue}{HTML}{000CC7}
\usepackage{comment}

\title{LegalLens: Leveraging LLMs for Legal Violation Identification in Unstructured Text}

\author{%
Dor Bernsohn$^{\textcolor{blue}{\star}}$\hspace{3mm}%
\textbf{Gil Semo}$^{\textcolor{blue}{\star}}$\hspace{3mm}%
\textbf{Yaron Vazana}$^{\textcolor{blue}{\star}}$\hspace{3mm}%
\textbf{Gila Hayat}$^{\textcolor{blue}{\star}}$\hspace{3mm}%
\textbf{Ben Hagag}$^{\textcolor{blue}{\star}}$ \\
\textbf{Joel Niklaus}$^{\textcolor{orange}{\dagger}}$\hspace{3mm}%
\textbf{Rohit Saha}$^{\textcolor{green}{\ddagger}}$\hspace{3mm}%
\textbf{Kyryl Truskovskyi}$^{\textcolor{green}{\ddagger}}$ \\
\\
$^{\textcolor{blue}{\star}}$Darrow AI Ltd., Tel Aviv, Israel \texttt{\{firstname.lastname\}@darrow.ai} \\
$^{\textcolor{orange}{\dagger}}$Niklaus.ai, Bern, Switzerland \texttt{joel@niklaus.ai} \\
$^{\textcolor{green}{\ddagger}}$Georgian.io, Toronto, Canada \texttt{\{firstname\}@georgian.io}
}
\begin{document}
\maketitle
\begin{abstract}

In this study, we focus on two main tasks, the first for detecting legal violations within unstructured textual data, and the second for associating these violations with potentially affected individuals. We constructed two datasets using Large Language Models (LLMs) which were subsequently validated by domain expert annotators. Both tasks were designed specifically for the context of class-action cases. The experimental design incorporated fine-tuning models from the BERT family and open-source LLMs, and conducting few-shot experiments using closed-source LLMs. Our results, with an F1-score of 62.69\% (violation identification) and 81.02\% (associating victims), show that our datasets and setups can be used for both tasks. Finally, we publicly release the datasets and the code used for the experiments in order to advance further research in the area of legal natural language processing (NLP).
\end{abstract}

\section{Introduction}

The widespread use of the internet has changed how information moves and connects in our society. Every day, the digital domain is flooded with a multitude of textual data, spanning from news articles and reviews to social media posts \footnote{\url{https://www.internetlivestats.com/total-number-of-websites}}. Within this sea of unstructured text, legal violations can often go unnoticed, concealed by the vast amount of surrounding information. These violations not only pose potential harm to individuals and entities but also challenge the very fabric of legal and ethical standards in the digital era. The significance of addressing these hidden violations cannot be overstated; as they have widespread implications for individual rights, societal norms, and the principles of justice. As a result, there is a pressing need to develop sophisticated methods to sift through the noise and identify these breaches.

\begin{figure}[ht]
    \centering
    \includegraphics[width=0.5\textwidth]{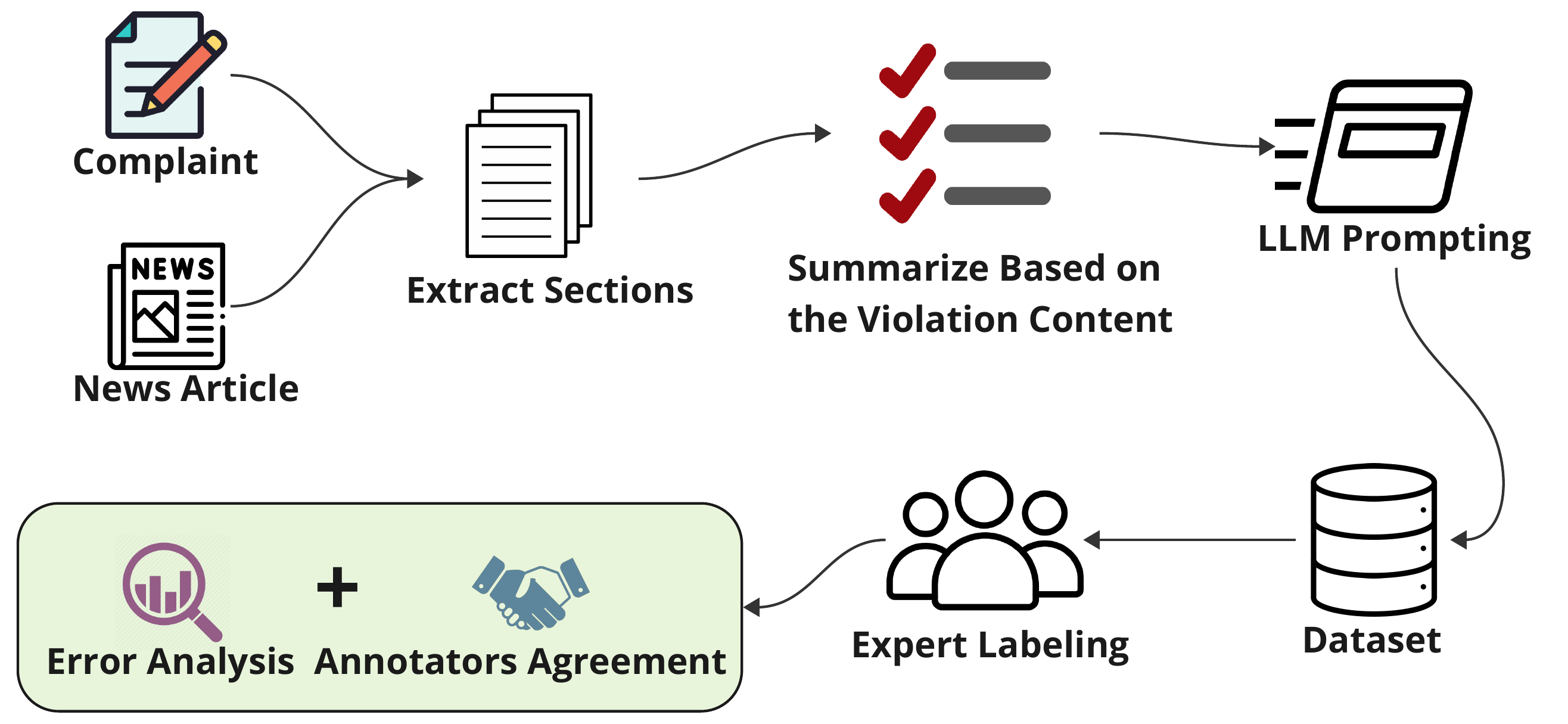}
    \caption{A visual representation of the data generation flow, illustrating the step-by-step process from raw input to the final synthesized dataset.}
    \label{fig:data_generation_flow}
\end{figure}

Legal violations often leave data trails. To detect these trails for pinpointing the violations, previous studies have often relied on specialized models tailored for specific domain applications \cite{9162683, 9206907}. These models, while effective in their specific domains, lack the versatility needed to address the wide array of legal violations that can occur across different contexts.

Legal violation identification aims to automatically uncover legal violations from unstructured text sources and assign potential victims to these violations.  We designed two setups, one for each task, the first for solving the legal violation identification task (a.k.a Identification Setup) using named entity recognition (NER), and the other for associating these violations with potentially affected individuals (a.k.a Resolution Setup) using natural language inference (NLI). Our dataset for the NER task is not limited to any specific domain,  while the NLI dataset is focused on four common legal domains. Followed by recent research in the field of data generation \cite{leiker2023prototyping, veselovsky2023generating, hamalainen2023evaluating}, we chose to employ GPT-4 \cite{openai2023gpt4} for synthetic data generation due to his ability to produce a large, diverse, and high-quality dataset that closely mimics the syntactic complexity of legal language, offering a scalable and ethically sound alternative to manual data crafting. We employed a thorough verification process to validate the data for both its realistic and complexity. Our approach involved automated data generation based on real-world event contexts in the English language, complemented by manual reviews conducted by seasoned legal annotators on the generated data.

\subsection*{Contributions}
The contributions of this paper are three-fold:
\begin{itemize}
    \item We introduce two dedicated datasets for legal violation identification, based on previous class action cases and legal news. These datasets, which include new legal entities, were generated using LLMs and validated by domain experts.
    \item We evaluate various language models, including BERT-based models and LLMs, across two different NLP tasks, offering valuable insights into their applicability and limitations in the context of legal NLP.
    \item We implement a two-setup approach employing both NER and NLI tasks, providing a methodology for legal violation detection and resolution.
\end{itemize}

\subsection*{Main Research Questions}
We believe numerous violations exist in unstructured text. Our aim is to uncover these violations and link them to relevant prior class actions. This study focuses on the following key research questions:\\
\textbf{RQ1: } To what extent do our newly introduced datasets enhance the performance of language models in identifying legal violations within unstructured text and associate victims to them?\\
\textbf{RQ2: } How effectively do the language models adapt to new, unseen data for the purpose of identifying legal violations and correlating them with past resolved cases across different legal domains?\\
\textbf{RQ3: } What is the level of difference between machine-generated and human-generated text in the context of legal violation identification?\\

\section{Related Work}

Previous works in the field of legal violation identification mostly focused on domain-specific topics, encompassing areas such as compliance, data privacy, and industry-specific regulations. For instance, \citet{amaral2023nlp} evaluates data agreements for compliance with European privacy laws using NLP techniques. \citet{9162683} used NER to identify personal information in datasets, thereby uncovering instances of online data privacy breaches. \citet{nyffenegger2023anonymity} used LLMs to attempt re-identification of anonymized persons from court decisions. Additionally, neural networks have been used to classify and annotate violation cases in specific industries like power supply \cite{9206907}. These studies, while valuable, have generally been limited to specific types of legal domains or particular sectors. Our work contributes to this existing body of research by introducing a dataset designed for broader applicability in identifying various types of legal violations.

Prior research has explored the use of Large Language Models (LLMs) for synthetic data generation \cite{rosenbaum2022clasp, rosenbaum2022linguist}, beneficial in situations with scarce authentic data \cite{brown2020language}. In fact, training models on synthetic data led to improved outcomes in benchmarks like SQUAD1.1 \cite{puri2020training}. However, human-curated data often provides a richness that is hard to replicate \cite{moller2023prompt, ding2022gpt}. In this paper, we present a multi-step validation method to discern between real-world and machine-generated content, addressing the inherent limitations of relying solely on synthetic data.

Previous studies indicate that LLMs are capable of explaining legal terms present in legislative documents by drafting explanations of how previous courts explained the meaning of statutory terms \cite{savelka2023explaining}.  Moreover, the models demonstrated analytical depth in court decision analysis, rivaling seasoned law students \cite{savelka2023can}. In this study, we created a dataset based on a previous lawsuits legislation background, rather than examining existing records.

While LLMs \cite{radford2019language} have been employed to enhance datasets for event detection tasks \cite{veyseh2021unleash}, our methodology advances this by generating pairs of specific violations and their corresponding events, using data from previously settled lawsuits. Unlike \citet{koreeda2021contractnli}, who concentrated on NLI in the context of legal contracts, our research introduces an NLI dataset based on class-action cases. Additionally, NER has been increasingly applied in the legal domain, including efforts to extract entities from Indian court judgments \cite{kalamkar2022named} and other legal texts \cite{luz2018lener, angelidis2018named, leitner2019fine}. Despite these advancements, existing research has largely focused on a standard set of entity types, such as parties (plaintiff and defendant), judges, court name and law/citation. Our work introduces a new set of entity types that have not been previously explored in legal NER research \cite{puaiș2021named, luz2018lener, dozier2010named, leitner2020dataset, skylaki2020named, kalamkar2022named}, thereby expanding the scope and applicability of NER in legal contexts.

\section{Curating Custom Legal Datasets: A Multi-stage Approach to NER and NLI Tasks}

Existing datasets may not adequately address the diverse range of legal violations and contexts central to our study, which is not in specific areas. To overcome these challenges, we employed a systematic and carefully planned data generation process, consisting of three stages: prompting, labeling, and data validation. This approach aimed at creating two robust datasets for two NLP tasks in the legal domain. We chose to focus on two key tasks: 
\begin{itemize}
\item NER (classifying tokens into predefined entities) for identifying violations. NER has been employed to define novel legal entities, enabling precise localization of pertinent information necessary for the extraction of legitimate legal violations, as detailed in Table \ref{table:ner_entities} in Appendix \ref{sec:data_destribution}.
\item NLI (classifying a hypothesis and a premise into entailed/contradict/neutral) for matching these violations with known, resolved class-action cases. NLI facilitates the correlation of multiple unstructured text associated with the same violation, thereby enabling the matching of extracted violations identified by the NER task with pre-existing legal complaints of class action cases. 
\end{itemize}

This dual-setup approach was designed to mimic the process of legal violation detection and resolution, generating high-quality data that closely resembles real-world scenarios.

Based on recent research in prompt-based methods \cite{liu2023pre}, our study employs prompts for a variety of reasons. LLMs have been shown to adapt to specialized tasks through techniques like instruction tuning \cite{wei2021finetuned}, reinforcement learning from human feedback \cite{ouyang2022training}, and in-context learning \cite{brown2020language} when prompted with natural language instructions. Prompts facilitate task-specific optimization, a quality emphasized by DialogPrompt \cite{gu2021response}, which aligns with our focus on NER and NLI in the legal domain by fine-tuning on the generated dataset. Additionally, the sensitivity of prompts in context, as demonstrated in Time-aware Prompts in Text Generation \cite{cao2022time}, is crucial for understanding specific legal contexts like resolved class-action cases. As a result, our methodology leverages a prompt-based approach, optimized for the legal domain, to generate high-quality data for NER and NLI tasks.

\subsection{Interconnection Between NER and NLI}

The process of identifying and resolving legal violations in unstructured text involves the collaborative use of NER and NLI. Initially, a NER model scans the text to detect 'VIOLATION' entities, and if a potential violation is tagged with a high-confidence score, it's considered for further analysis. Subsequently, the text is processed through an NLI model in a pair-wise fashion against a dataset of closed settlements. If the NLI model finds a logical entailment between the text and any of the settled cases, indicating a substantial similarity, the corresponding complaints are flagged as candidates for matching with the specific user's complaint, potentially qualifying them for inclusion in a settlement fund. This streamlined approach harnesses the strengths of both NER and NLI to efficiently identify and associate potential legal violations with relevant precedents.

\subsection{NER Data Generation}

NER can be framed as a token classification task, wherein, the objective is to classify each word in a sentence as an entity class. In our dataset, there are four such entities; \textit{Law}, \textit{Violation}, \textit{Violated By}, and \textit{Violated On}.

For the NER task, our foundational data source was class action complaints, as described in \cite{semo2022classactionprediction}. A complaint, often referred to as a plaintiff's plea, is a formal legal document that initiates a lawsuit. It outlines the complaints of the plaintiff and specifies the relief sought from the court. From each of these complaints, we extracted relevant sections such as \textit{allegations}, \textit{counts}, and \textit{legal arguments} that were pertinent to our study, ensuring relevance and precision. These sections encapsulate the main context of the alleged violations. They were subsequently summarized through the utilization of GPT-4 \cite{openai2023gpt4} to capture the core essence of the violation content, and were employed as the context in the subsequent prompts.

For a visual representation of our data generation process, refer to Figure \ref{fig:data_generation_flow}.

\subsection*{Prompt}
For the NER task, we devised two unique prompting strategies: explicit and implicit. The explicit method not only emphasizes the inclusion of multiple distinct entities but also underscores the specific order of their appearance, adding a layer of complexity and structure to the generated content (refer to figure \ref{figure:NER_datagen_prompts} in the Appendix). This approach ensures that the content is not only diverse but also adheres to certain structural guidelines, which contain task descriptions, specific instructions, and few-shot examples. Conversely, the implicit strategy focuses solely on a singular entity, specifically the content that describes the violation, refer to figure \ref{figure:NER_datagen_prompts} in the Appendix.

Furthermore, both strategies incorporate additional parameters such as the cause of action, industry, and context. The inclusion of these parameters refines the generated content, tailoring it to specific scenarios and ensuring its relevance to the desired domain. By employing the explicit approach, we capture the comprehensive nature of a scenario, whereas the implicit method provides a concise perspective on one specific aspect.

\subsection{NLI Data Generation}

NLI can be framed as a classification task, wherein, the objective is to compare a premise to a hypothesis, and predict one of the three classes:  (1) \textit{Entailment} - where the hypothesis is contained and can be supported by the premise, (2) \textit{Contradiction} - when the hypothesis contradicts the premise, (3) \textit{Neutral} - when the premise neither entails nor contradicts the hypothesis.

For the NLI task, our data source consisted articles taken from a legal news website. Each news article was first summarized, by prompting GPT-4 \cite{openai2023gpt4}, to capture its legal grounds. By summarizing, we ensured that the data was concise yet comprehensive by keeping only the legal violation section and removing background parts. This summarized content served as the premise. Using this premise, the model was tasked to generate a hypothesis that mimicked real-world scenarios. The intention behind this design was to create diverse records that spanned various legal areas. Table \ref{table:NLI_DOMAINS} in Appendix \ref{sec:data_destribution} presents the NLI data distributions.

\subsection*{Prompt}

In this setup, we aimed to create scenarios that mirror real-life accounts of potential violations. We generated texts that mimic common situations where individuals share concerns, like online reviews or social media posts. The goal was to produce narratives that implicitly describe the effects of a violation. We added variations in attributes such as the writers age and gender and the text format to capture a wide range of experiences.

\begin{table*}[ht]
\scriptsize
\centering
\caption{Comparison of different methodologies for NER. The table showcases various models, their sizes, and the method employed, along with their performance metrics.}

\begin{tabular}{lrccccc} 
\toprule
Model & Size & Method & F1 & Precision & Recall \\
\midrule
nlpaueb/legal-bert-small-uncased  & 35M & Fine-tune & $48.90_{\pm 0.39}$ & $41.92_{\pm 0.80}$ & $58.69_{\pm 0.52}$  \\
distilbert-base-uncased  & 66M & Fine-tune & $49.71_{\pm 0.83}$ & $42.19_{\pm 0.89}$ & $60.50_{\pm 0.77}$ \\
\midrule
bert-base-cased  & 108M & Fine-tune & $54.80_{\pm 0.64}$ & $47.23_{\pm 1.06}$ & $65.28_{\pm 1.01}$  \\
bert-base-uncased  & 109M & Fine-tune & $53.22_{\pm 1.42}$ & $45.86_{\pm 1.68}$ & $63.42_{\pm 1.11}$  \\
roberta-base & 125M  & Fine-tune & \textbf{62.69}$_{\pm \textbf{0.69}}$ & $56.58_{\pm 1.12}$ & \textbf{70.30}$_{\pm \textbf{0.73}}$ \\
nlpaueb/legal-bert-base-uncased  & 109M  & Fine-tune & $57.50_{\pm 0.94}$ & $50.34_{\pm 1.26}$ & $67.04_{\pm 0.71}$  \\
lexlms/legal-roberta-base  & 124M & Fine-tune & $59.73_{\pm 2.03}$ & $53.11_{\pm 2.27}$ & $68.25_{\pm 1.86}$  \\
joelito-legal-english-roberta-base & 124M & Fine-tune & $59.01_{\pm 1.74}$ & $52.52_{\pm 2.52}$ & $67.40_{\pm 0.85}$ \\
lexlms/legal-longformer-base  & 148M & Fine-tune & $62.30_{\pm 1.76}$ & \textbf{56.78}$_{\pm \textbf{2.14}}$ & $69.04_{\pm 1.32}$ \\
\midrule
lexlms/legal-roberta-large & 355M & Fine-tune & $50.23_{\pm 28.1}$ & $46.07_{\pm 25.8}$ & $55.22_{\pm 30.8}$ \\
lexlms/legal-longformer-large & 434M & Fine-tune & $37.63_{\pm 34.4}$ & $34.26_{\pm 31.3}$ & $41.76_{\pm 38.1}$ \\
joelito-legal-english-roberta-large & 355M & Fine-tune & $58.92_{\pm 4.28}$ & $52.88_{\pm 4.95}$ & $66.59_{\pm 3.22}$ \\
\midrule
Falcon  & 7B & QLoRA & $1.00_{\pm 0.50}$ & $39.50_{\pm 16.8}$ & $0.50_{\pm 0.20}$  \\
Llama-2 & 7B & QLoRA & $16.3_{\pm 4.10}$ & $34.10_{\pm 11.1}$ & $11.20_{\pm 2.60}$  \\
\midrule
OpenAI GPT-3.5 & 175B & Few-shot & $2.77_{\pm 0.12}$ & $1.78_{\pm 0.08}$ & $6.23_{\pm 0.29}$  \\
OpenAI GPT-4 & - & Few-shot & $13.55_{\pm 0.54}$ & $8.29_{\pm 0.37}$ & $37.1_{\pm 0.99}$  \\
\bottomrule
\label{table:model-performance-ner}
\end{tabular}
\end{table*}




\begin{table}[ht]
\scriptsize
\centering
\caption{Entity-specific F1 score for the best-performing NER model, `roberta-base`.}

\begin{tabular}{lcccc} 
\toprule
LAW & VIOLATION & VIOLATED BY & VIOLATED ON\\
\midrule
$77.57_{\pm 1.35}$ & $59.06_{\pm 0.55}$ & $76.88_{\pm 2.06}$ & $62.83_{\pm 2.57}$ \\
\bottomrule
\end{tabular}
\label{table:best-model-performance-NLI}
\end{table}

\section{Human Expert Annotations}

Data validation holds particular importance in our study due to the synthetic nature of the dataset. To ensure that the dataset is both realistic and challenging, we have implemented several validation methods. In this structured process, summaries of complaint documents and tasks for the NER and NLI models were generated automatically. Legal experts then carefully examined these auto-generated summaries and tasks. Their primary role was to meticulously review each output, ensuring that the summaries accurately reflected the key points of the complaints and that the tasks were correctly aligned with the context provided by these summaries. Additionally, each record was subjected to examination by several annotators, which serves to reduce potential bias in the evaluation. These annotators were tasked with identifying and suggesting any missing entities, as well as in checking for hallucinations—instances where the generated content might stray from factual accuracy. To maintain a rigorous and unbiased validation, all annotators received identical instructions, and the data presented to them was systematically shuffled. Their detailed examination was crucial in pinpointing discrepancies, unclear areas, or potential inaccuracies in both the summaries and the associated tasks. This thorough validation process, attentive to both content accuracy and the prevention of hallucinations and bias through multiple annotators review, ensures the integrity and quality of our synthetic dataset. Figure \ref{fig:annotation_plattform} in Appendix \ref{sec:annotation} presents a screenshot of the annotation platform we used.

Upon further examination of our data, a comparison between machine-generated and human-authored content revealed significant similarities. This comparison involved analyzing various linguistic and structural features of the texts. Both displayed identical average sentence lengths. Moreover, there was not significant difference between the character count between the generated content and the human-authored text. Additionally, when comparing the POS tags between the real text and the generated text, by averaging the total counts of each tag occurrences, the average difference was found to be 26\% and the median was 16\%.

A key part of our validation process was the classification task. In this task, three independent annotators had to distinguish between machine-generated and human-written records, a challenge also noted in recent research \cite{mitchell2023detectgpt, kirchenbauer2023watermark}. Our annotators' goal was to label each record based on its origin: machine-generated or human-written. The annotators achieved an average F1-score of 44.86\%. However, their Cohen's Kappa scores, which were 0.0821, 0.2149, and 0.0988, showed only minor agreement among them. This low level of agreement, as indicated by Cohen's Kappa scores, points out the complexity of the task. It also suggests that our machine-generated content closely resembled human writing, making it difficult even for experts to tell them apart. The use of Cohen's Kappa in our study is supported by its well-known effectiveness in binary classification tasks, especially in data annotation scenarios \cite{wang2019simplified}.

\section{Experiments}

In this section, we explore several methods to tackle the challenging and realistic setups that we created. More precisely, we analyzed the performance of language models on these setups by conducting three sets of experiments. (1) We evaluated models that are inspired by the BERT architecture through the process of fine-tuning~\cite{sun2020finetune}. (2) We explored LLMs such as Falcon-7B, Llama-2-7B and Llama-2-13B through the process of parameter efficient fine-tuning \cite{houlsby2019parameterefficient, hu2021lora}. (3) Thanks to their out-of-the-box generalization capabilities, we assessed OpenAI's GPT-3.5~\cite{brown2020language} and GPT-4~\cite{openai2023gpt4} models.

\subsection{Setup}

\paragraph{NER}
Our dataset is categorized by Cause of Action (CoA). CoA refers to a set of facts or legal reasons that justify the right to sue or seek legal remedy in a court of law. Due to the potential overlap and similarities between different CoAs, there's a risk of data leakage when training models. To mitigate this, we adopted a strategy where CoAs present in the training set were excluded from the test set. This ensures that the model is evaluated on entirely distinct CoAs, preventing any inadvertent training on test data.

\paragraph{NLI}
Our dataset contains news articles across four legal domains. Given the similarities in the legal merits between these domains, there is a potential risk of data leakage related to the legal attributes of the cases. To address this issue, we employed a leave-one-out approach. In this method, we tested each legal domain separately while training the model on the other domains. This 'leave-one-out' method strengthens the model's ability to generalize by ensuring it is evaluated on entirely unseen data, reducing the risk of overfitting by its small size. By exposing the model to a variety of legal domains during training, but withholding one domain for testing, we mimic real-world scenarios where the model will encounter previously unseen data.

\subsection{Model Classes}

\paragraph{BERT Models}

In this setting, we assess the effectiveness of transformer-based language models~\cite{vaswani2017}. We fine-tuned RoBERTa~\cite{liu2019roberta}, DistilBERT~\cite{sanh2019distilbert} and BERT~\cite{devlin2018bert} models. Additionally, we evaluated their legal counterparts, i.e., Legal-BERT~\cite{chalkidis2020legalbert} and Legal-RoBERTa~\cite{chalkidis-garneau-etal-2023-lexlms}. Furthermore, we evaluated models~\cite{mamakas2022processing} based on the Longformer architecture~\cite{beltagy2020longformer}. Following this, we also assessed the Legal-English-RoBERTa models, which are specialized versions tailored for legal English~\cite{niklaus2023multilegalpile}. We utilized the AutoModel family classes from the HuggingFace Transformers library to train the models. Each model was trained for 10 epochs with an initial learning rate of $2e-5$. In addition, we used early-stopping to prevent overfitting.

\paragraph{Open-Source LLMs}

In this setting, we evaluated Falcon~\cite{falcon40b} and Llama2s~\cite{touvron2023llama} performance. More precisely, we considered the 7 billion parametric version of Falcon, and 7 and 13 billion versions of Llama2. Following the success of Parameter Efficient Fine-Tuning methodologies for fine-tuning LLMs, we leveraged QLoRA~\cite{dettmers2023qlora} due to its superior performance over other methods. Figure~\ref{figure:llm-prompts} shows the prompt that we designed to guide the tuning process.

The prompt has two parts: Input and Output. The Input contains the sentence on which NER and NLI have to be performed. The Output contains the format in which the LLM has to predict the entities contained in the sentence. It is important to note that during inference, we prompt the model to generate the required output by only including the Input section.

We employed HuggingFace's AutoModelForCausalLM class for fine-tuning, available under an Apache-2.0 license\footnote{\url{https://github.com/huggingface/transformers}}. Each model underwent training for 20 epochs with an initial learning rate of 2e-4, a QLoRA rank of 64, and a dropout rate of 0.25. We used this configuration across both NER and NLI tasks.

\paragraph{Closed-Source LLMs}
We evaluate OpenAI's GPT-4~\cite{openai2023gpt4} and OpenAI's GPT-3.5~\cite{brown2020language} models for few-shot NER and NLI without any fine-tuning, using the matching production models of August 2023. We use the Langchain\footnote{\url{https://github.com/langchain-ai/langchain}} client, available under an Apache-2.0 license, with few-shot prompts, as demonstrated in Figure~\ref{figure:few-shot_prompts}.
In all experiments, we set the temperature to 0.7 and used 9 random samples from the training dataset as few-shot examples. We employed the same prompts as those used for open-source models and the same evaluation mechanism. Each API call was repeated five times.

\begin{table*}[ht]
\scriptsize
\centering
\caption{Macro F1 evaluation of various model architectures for the NLI task across different legal entities.}

\begin{tabular}{lcccc} 
\toprule
Model & Consumer Protection & Privacy & TCPA & Wage \\
\midrule
nlpaueb-legal-bert-small-uncased & $60.8_{ \pm 7.1}$ & $49.6_{ \pm 14.}$ & $47.6_{ \pm 11.}$ & $56.7_{ \pm 6.0}$ \\
distilbert-base-uncased & $79.8_{ \pm 2.0}$ & $53.9_{ \pm 13.}$ & $72.1_{ \pm 9.3}$ & $71.2_{ \pm 7.3}$ \\
\midrule
bert-base-cased & $65.5_{ \pm 9.2}$ & $39.9_{ \pm 18.}$ & $58.9_{ \pm 16.}$ & $65.5_{ \pm 13.}$ \\
bert-base-uncased & $69.3_{ \pm 7.7}$ & $36.3_{ \pm 16.}$ & $69.5_{ \pm 7.2}$ & $64.0_{ \pm 16.}$ \\
roberta-base & $82.9_{ \pm 4.5}$ & $62.0_{ \pm 5.0}$ & $69.5_{ \pm 31.}$ & $69.7_{ \pm 29.}$ \\
lexlms-legal-roberta-base & $45.8_{ \pm 5.8}$ & $27.3_{ \pm 7.9}$ & $48.6_{ \pm 14.}$ & $44.4_{ \pm 19.}$ \\
joelito-legal-english-roberta-base & $61.6_{ \pm 14.2}$ & $33.1_{ \pm 12.2}$ & $55.8_{ \pm 9.95}$ & $48.6_{ \pm 17.9}$ \\
lexlms-legal-longformer-base & $58.3_{ \pm 16.}$ & $27.8_{ \pm 4.6}$ & $54.8_{ \pm 11.}$ & $54.5_{ \pm 11.}$ \\
\midrule
lexlms-legal-roberta-large & $18.1_{ \pm 0.7}$ & $20.2_{ \pm 8.1}$ & $15.3_{ \pm 1.8}$ & $16.6_{ \pm 0.0}$ \\
lexlms-legal-longformer-large & $19.2_{ \pm 1.3}$ & $17.5_{ \pm 0.6}$ & $25.5_{ \pm 24.}$ & $26.3_{ \pm 21.}$ \\
joelito-legal-english-roberta-large & $16.4_{ \pm 3.3}$ & $20.2_{ \pm 5.8}$ & $47.3_{ \pm 30.3}$ & $27.3_{ \pm 23.9}$ \\
\midrule
Falcon 7B & \textbf{87.2}$_{\pm \textbf{3.1}}$ & \textbf{84.5}$_{\pm \textbf{8.8}}$ & \textbf{83.9}$_{\pm \textbf{0.9}}$ & 68.5 $_{\pm 11.}$ \\

Llama-2 7B & $47.2_{ \pm 5.9}$ & $47.8_{ \pm 10.}$ & $63.5_{ \pm 7.3}$ & $63.7_{ \pm 14.}$ \\
Llama-2 13B & $63.1_{ \pm 8.0}$ & $75.2_{ \pm 6.5}$ & $63.9_{ \pm 10.}$ & \textbf{86.5}$_{ \pm \textbf{5.6}}$ \\
\midrule
OpenAI GPT-3.5 & $17.8_{ \pm 2.6}$ & $18.12_{ \pm 3.1}$ & $15.09_{ \pm 1.9}$ & $12.91_{ \pm 5.4}$ \\
OpenAI GPT-4 & $49.83_{ \pm 19.}$ & $48.44_{ \pm 9.4}$ & $37.04_{ \pm 7.4}$ & $52.48_{ \pm 11.6}$ \\
\bottomrule
\end{tabular}
\label{table:model-performance-NLI}
\end{table*}

\section{Results}

\subsection{NER}

Table~\ref{table:model-performance-ner} presents the performance metrics of various models. Interestingly, BERT-based models with fewer parameters outperform LLMs by a significant margin. This disparity in performance is due to the difference in objective functions that the different model classes use. BERT-based models employ the cross-entropy objective function per token, providing a stronger gradient signal. Furthermore, the label space is well constrained by the number of possible entities in our data set. On the other hand, LLMs have been fine-tuned via causal language modeling, wherein the task is to learn the joint probability distribution of all tokens by maximizing the likelihood of the data. The gradient signal in the case of fine-tuning LLMs is not as fine-grained as cross-entropy. This is because the label space, i.e., the number of possibilities to predict the next token from, far exceeds the number of required entities.

Across BERT-based models, we notice interesting trends. First, \textit{roberta-base} model attains the best performances, achieving an F1 score of $62.69\%$ and Recall of $70.3\%$. Second, the performance across all metrics improved as model complexity grew, except for Longformer-based models and joelito-legal-english-roberta-based models.

Focusing on LLMs, we observed that both open-source and close-source models perform poorly on this task. Closer analysis of predictions indicated incorrect B-token prediction in generated text.  These errors were propagated to the next predictions, causing the LLMs to misclassify the tokens and place them into incorrect entities.

\subsection{NLI}

Table~\ref{table:model-performance-NLI} shows domain-specific performances across all model classes. In contrary to trends discovered in the NER experiments, in NLI we noticed that LLMs outperform BERT-based models by a very significant margin. Unlike NER, in NLI, LLMs are fine-tuned to predict only one token, i.e., either of \textit{entailed}, \textit{contradict}, and \textit{neutral}. Additionally, the NLI task had only 312 samples, and LLMs learn relatively better in low data situations and generalize well to out-of-distribution (OOD) test data sets \cite{brown2020language}.

Except for domain \textit{Wage}, \textit{Falcon 7B} achieved the highest performance across domains (\textit{Consumer Protection}, \textit{Privacy}, and \textit{TCPA}). \textit{Falcon 7B} attained the highest Macro F1 metric, demonstrating its OOD capabilities. Among BERT-based models, \textit{roberta-base} once again achieved the best performance, similar to NER tasks.

\section{Error Analysis}
To improve our models and enrich our understanding, we conducted a thorough error analysis of top-performing models across tasks. This analysis identifies their limitations, providing a clear roadmap for future refinements.

\subsection{NER}
In evaluating our NER model, the entity type "VIOLATION" exhibited the lowest F1 score. This entity is often lengthy and contextually complex, making it a challenging target for accurate identification. We conducted an error analysis on a subset of hard cases to understand the model's limitations.

The errors fall into three categories: truncation errors, context misunderstanding, and incorrect entity identification. For instance, in the sentence \textit{"I've been getting these [VIOLATION] constant calls on my cell phone from some company that won't quit [VIOLATION]."}, the model predicted \textit{"constant calls on"} instead of the actual entity. This truncation error suggests the model captures only the initial segment but fails to include the entire scope. In another example, \textit{"They've been [VIOLATION] failing to disclose that their educational programs were underperforming [VIOLATION]."}, the model predicted \textit{"disclose"}, indicating a context misunderstanding. Notably, when the model completely misses the target, it often predicts a much shorter entity, suggesting a bias towards shorter answers when uncertain.

The model struggles with the "VIOLATION" entity type, particularly with longer and more complex entities. Fine-tuning the model with a diverse, context-rich training set could improve its performance. Existing literature also suggests that NER models often struggle with complex entities \cite{dai2018recognizing}, underscoring the need for continued research in this area.

\begin{figure}[ht]
    \centering
    \includegraphics[width=0.5\textwidth]{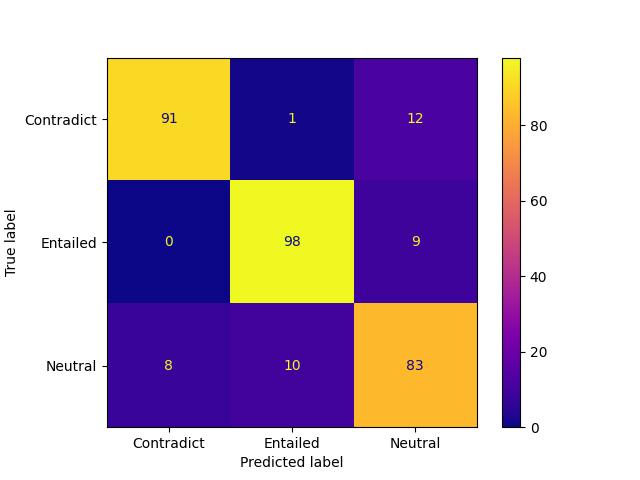}
    \caption{NLI Confusion Matrix derived from the top performer model (Falcon 7B's) predictions.}
    \label{fig:NLI Confusion Matrix}
\vspace{-5mm}
\end{figure}

\subsection{NLI}
In the error analysis of our best performing NLI model, Falcon 7B, we consolidated the model errors across different legal domains to form a comprehensive view. Our focus was on two types of classification errors: first-class errors, which involve confusions between "Contradict" and "Entailed", and second-class errors, which are misclassifications of "Contradict" or "Entailed" as "Neutral". Figure~\ref{fig:NLI Confusion Matrix} shows that while Falcon 7B performs well in avoiding first-class errors, it exhibits a substantial number of second-class errors. The high rate of such errors indicates that the model finds it challenging to handle more nuanced cases where it is difficult to discern whether the person was affected by the violation or not.

Although Falcon 7B outperforms other models in this task, it strugglesin accurately classifying statements related to wage areas. This could be attributed to the complexities and ambiguities of the wage norms, which make it challenging to clearly determine whether a wage violation has occurred. Therefore, investigating different token lengths to provide more context or fine-tuning the model to better navigate these intricate wage scenarios could be valuable directions for future work.

\section{Conclusions and Future Work}

\subsection{Answers to the Research Questions}
\noindent \textbf{RQ1}: \emph{To what extent do our newly introduced datasets enhance the performance of language models in identifying legal violations within unstructured text and associate victims to them?}
The study introduced new entities in the datasets. This addition improved the ability of language models to identify legal violations in unstructured text. With these new entities, the roberta-base model achieved an F1-score of 62.69\% in identifying violations and 81.02\% (Falcon 7B model) in linking them to victims. This demonstrates that our new approach, which focuses on identifying and associating violations to victims, has been successful, yet there remains potential for further refinements and improvements.

\noindent \textbf{RQ2}: \emph{How effectively do the language models adapt to new, unseen data for the purpose of identifying legal violations and correlating them with past resolved cases across different legal domains?}
Our experiments assessed language models' adaptability to unseen data, especially in the context of identifying legal violations and correlating them with past resolved cases across different legal domains. While BERT-based models demonstrated strong performance in certain tasks, LLMs like Falcon-7B excelled in low-data scenarios, particularly in associating violations with resolved cases. This suggests that these models effectively adapt to new data, especially when the data is limited.

\noindent \textbf{RQ3}: \emph{What is the level of difference between machine-generated and human-generated text in the context of legal violation identification?}
Our validation process involved a comparison between machine-generated and human-authored content. The findings revealed that the two types of content were strikingly similar in terms of average sentence lengths and character count. When expert annotators were tasked to distinguish between machine-generated and human-written records, they achieved an average F1-score of 44.86\%. The low level of agreement among the annotators indicates that our machine-generated content closely resembles human writing, making it challenging even for experts to differentiate between the two.

\subsection{Conclusion}
In this study, by leveraging LLMs and expert validation, we introduced a dual setup approach to identify legal violations from text. Our approach uses (1) NER to pinpoint violations, resulting in an F1-score of 62. 69\% and (2) NLI to associate these violations with resolved cases, resulting in an F1-score of 81.02\%. We created two specialized datasets to advance research in this field.

\subsection{Future Work}

\paragraph{Expanding Legal Areas}
In future iterations, we aim to expand the dataset to include a broader range of legal areas. By incorporating diverse legal texts, we hope to create a more representative dataset for legal violation identification.

\paragraph{Incorporating Multiple Jurisdictions}
While our current dataset is heavily focused on common law in US courts, future work will aim to integrate legal texts from various global jurisdictions, including civil law systems. This will not only enhance the datasets diversity but also improve the robustness and applicability of models trained on it.

\paragraph{Fact Matching}
An avenue for future work is the integration of fact matching. Developing algorithms for cross-referencing facts across sources can enhance the accuracy of legal violation identification, especially when a single source might not provide a complete picture. \cite{thorne2018fever, jiang2020hover}

\section*{Limitations}
\paragraph{Focus on Common Law in US Courts}
A primary limitation of our dataset is its focus on US common law. While this deepens understanding of US legal principles and precedents, it may not apply to civil law jurisdictions or non-US legal systems. The nuances, interpretations, and applications of laws can vary significantly across different jurisdictions, and our dataset, being US-centric, might not capture these variations adequately.

\paragraph{Coverage of Areas of Law}
While our dataset provides a comprehensive overview of legal violations from various text sources, it does have its limitations in terms of the breadth of legal areas covered. The current dataset predominantly focuses on specific areas of law, potentially overlooking nuances and intricacies of other legal domains. For instance, while we have extensively covered areas like consumer protection and privacy, other equally significant areas such as intellectual property, environmental law, or international law might not have been represented with the same depth.

\section*{Ethics Statement}
The primary objective of this research is to revolutionize the identification and understanding of legal violations within the sprawling landscape of online text. By introducing a novel dataset specifically tailored for Named Entity Recognition (NER) and Natural Language Inference (NLI) tasks in the legal context, we aim to significantly advance the field of Natural Language Processing (NLP) and its applications in law. Our research holds the potential to greatly assist legal professionals in efficiently identifying and addressing legal violations, thereby contributing to a safer and more equitable digital society.

In the pursuit of this objective, we have employed LLMs, specifically GPT-4~\cite{openai2023gpt4}, for data generation, and have subjected the generated data to rigorous validation by expert annotators. This dual-layered approach ensures the quality and reliability of our dataset, while also providing a comprehensive range of examples that can be generalized across various domains.

However, we acknowledge that the deployment of machine learning models in the legal domain is fraught with ethical considerations \cite{tsarapatsanis2021ethical}. Automating the detection of legal violations could inadvertently lead to false positives or negatives, with serious implications for individual rights and the rule of law. Therefore, we stress that our technology is intended to serve as a supplementary tool for legal professionals, rather than a replacement. It is essential that any application of our dataset and subsequent models be conducted responsibly with a thorough understanding of the limitations and biases that may be inherent in automated systems.

Moreover, we recognize the ethical imperative of data privacy and confidentiality, especially given the sensitive nature of legal texts. All data used in this research have been anonymized and stripped of personally identifiable information to the best of our ability, in compliance with relevant data protection regulations. All the data utilized in this study is sourced from publicly accessible online platforms and does not infringe on any individuals or entities proprietary rights.

\section*{Acknowledgements}
We express our gratitude to the annotation group at Darrow.ai for their contributions to the dataset.

\bibliography{anthology,custom}

\begin{thebibliography}{55}
\expandafter\ifx\csname natexlab\endcsname\relax\def\natexlab#1{#1}\fi

\bibitem[{Almazrouei et~al.(2023)Almazrouei, Alobeidli, Alshamsi, Cappelli, Cojocaru, Debbah, Goffinet, Heslow, Launay, Malartic, Noune, Pannier, and Penedo}]{falcon40b}
Ebtesam Almazrouei, Hamza Alobeidli, Abdulaziz Alshamsi, Alessandro Cappelli, Ruxandra Cojocaru, Merouane Debbah, Etienne Goffinet, Daniel Heslow, Julien Launay, Quentin Malartic, Badreddine Noune, Baptiste Pannier, and Guilherme Penedo. 2023.
\newblock {Falcon-40B}: an open large language model with state-of-the-art performance.

\bibitem[{Amaral et~al.(2023)Amaral, Azeem, Abualhaija, and Briand}]{amaral2023nlp}
Orlando Amaral, Muhammad~Ilyas Azeem, Sallam Abualhaija, and Lionel~C Briand. 2023.
\newblock Nlp-based automated compliance checking of data processing agreements against gdpr.
\newblock \emph{IEEE Transactions on Software Engineering}.

\bibitem[{Angelidis et~al.(2018)Angelidis, Chalkidis, and Koubarakis}]{angelidis2018named}
Iosif Angelidis, Ilias Chalkidis, and Manolis Koubarakis. 2018.
\newblock Named entity recognition, linking and generation for greek legislation.
\newblock In \emph{JURIX}, pages 1--10.

\bibitem[{Beltagy et~al.(2020)Beltagy, Peters, and Cohan}]{beltagy2020longformer}
Iz~Beltagy, Matthew~E. Peters, and Arman Cohan. 2020.
\newblock \href {http://arxiv.org/abs/2004.05150} {Longformer: The long-document transformer}.
\newblock \emph{CoRR}, abs/2004.05150.

\bibitem[{Brown et~al.(2020)Brown, Mann, Ryder, Subbiah, Kaplan, Dhariwal, Neelakantan, Shyam, Sastry, Askell et~al.}]{brown2020language}
Tom Brown, Benjamin Mann, Nick Ryder, Melanie Subbiah, Jared~D Kaplan, Prafulla Dhariwal, Arvind Neelakantan, Pranav Shyam, Girish Sastry, Amanda Askell, et~al. 2020.
\newblock Language models are few-shot learners.
\newblock \emph{Advances in neural information processing systems}, 33:1877--1901.

\bibitem[{Cao and Wang(2022)}]{cao2022time}
Shuyang Cao and Lu~Wang. 2022.
\newblock Time-aware prompting for text generation.
\newblock \emph{arXiv preprint arXiv:2211.02162}.

\bibitem[{Chalkidis et~al.(2020)Chalkidis, Fergadiotis, Malakasiotis, Aletras, and Androutsopoulos}]{chalkidis2020legalbert}
Ilias Chalkidis, Manos Fergadiotis, Prodromos Malakasiotis, Nikolaos Aletras, and Ion Androutsopoulos. 2020.
\newblock \href {http://arxiv.org/abs/2010.02559} {{LEGAL-BERT:} the muppets straight out of law school}.
\newblock \emph{CoRR}, abs/2010.02559.

\bibitem[{Chalkidis* et~al.(2023)Chalkidis*, Garneau*, Goanta, Katz, and Søgaard}]{chalkidis-garneau-etal-2023-lexlms}
Ilias Chalkidis*, Nicolas Garneau*, Catalina Goanta, Daniel~Martin Katz, and Anders Søgaard. 2023.
\newblock \href {https://arxiv.org/abs/2305.07507} {{LeXFiles and LegalLAMA: Facilitating English Multinational Legal Language Model Development}}.
\newblock In \emph{Proceedings of the 61st Annual Meeting of the Association for Computational Linguistics}, Toronto, Canada. Association for Computational Linguistics.

\bibitem[{Dai(2018)}]{dai2018recognizing}
Xiang Dai. 2018.
\newblock Recognizing complex entity mentions: A review and future directions.
\newblock In \emph{Proceedings of ACL 2018, Student Research Workshop}, pages 37--44.

\bibitem[{Dettmers et~al.(2023)Dettmers, Pagnoni, Holtzman, and Zettlemoyer}]{dettmers2023qlora}
Tim Dettmers, Artidoro Pagnoni, Ari Holtzman, and Luke Zettlemoyer. 2023.
\newblock \href {http://arxiv.org/abs/2305.14314} {Qlora: Efficient finetuning of quantized llms}.

\bibitem[{Devlin et~al.(2018)Devlin, Chang, Lee, and Toutanova}]{devlin2018bert}
Jacob Devlin, Ming{-}Wei Chang, Kenton Lee, and Kristina Toutanova. 2018.
\newblock \href {http://arxiv.org/abs/1810.04805} {{BERT:} pre-training of deep bidirectional transformers for language understanding}.
\newblock \emph{CoRR}, abs/1810.04805.

\bibitem[{Ding et~al.(2022)Ding, Qin, Liu, Bing, Joty, and Li}]{ding2022gpt}
Bosheng Ding, Chengwei Qin, Linlin Liu, Lidong Bing, Shafiq Joty, and Boyang Li. 2022.
\newblock Is gpt-3 a good data annotator?
\newblock \emph{arXiv preprint arXiv:2212.10450}.

\bibitem[{Dozier et~al.(2010)Dozier, Kondadadi, Light, Vachher, Veeramachaneni, and Wudali}]{dozier2010named}
Christopher Dozier, Ravikumar Kondadadi, Marc Light, Arun Vachher, Sriharsha Veeramachaneni, and Ramdev Wudali. 2010.
\newblock \emph{Named entity recognition and resolution in legal text}.
\newblock Springer.

\bibitem[{Gu et~al.(2021)Gu, Yoo, and Lee}]{gu2021response}
Xiaodong Gu, Kang~Min Yoo, and Sang-Woo Lee. 2021.
\newblock Response generation with context-aware prompt learning.
\newblock \emph{arXiv preprint arXiv:2111.02643}.

\bibitem[{H{\"a}m{\"a}l{\"a}inen et~al.(2023)H{\"a}m{\"a}l{\"a}inen, Tavast, and Kunnari}]{hamalainen2023evaluating}
Perttu H{\"a}m{\"a}l{\"a}inen, Mikke Tavast, and Anton Kunnari. 2023.
\newblock Evaluating large language models in generating synthetic hci research data: a case study.
\newblock In \emph{Proceedings of the 2023 CHI Conference on Human Factors in Computing Systems}, pages 1--19.

\bibitem[{Houlsby et~al.(2019)Houlsby, Giurgiu, Jastrzebski, Morrone, de~Laroussilhe, Gesmundo, Attariyan, and Gelly}]{houlsby2019parameterefficient}
Neil Houlsby, Andrei Giurgiu, Stanislaw Jastrzebski, Bruna Morrone, Quentin de~Laroussilhe, Andrea Gesmundo, Mona Attariyan, and Sylvain Gelly. 2019.
\newblock \href {http://arxiv.org/abs/1902.00751} {Parameter-efficient transfer learning for nlp}.

\bibitem[{Hu et~al.(2021)Hu, Shen, Wallis, Allen-Zhu, Li, Wang, Wang, and Chen}]{hu2021lora}
Edward~J. Hu, Yelong Shen, Phillip Wallis, Zeyuan Allen-Zhu, Yuanzhi Li, Shean Wang, Lu~Wang, and Weizhu Chen. 2021.
\newblock \href {http://arxiv.org/abs/2106.09685} {Lora: Low-rank adaptation of large language models}.

\bibitem[{Jiang et~al.(2020)Jiang, Bordia, Zhong, Dognin, Singh, and Bansal}]{jiang2020hover}
Yichen Jiang, Shikha Bordia, Zheng Zhong, Charles Dognin, Maneesh Singh, and Mohit Bansal. 2020.
\newblock Hover: A dataset for many-hop fact extraction and claim verification.
\newblock In \emph{Findings of the Association for Computational Linguistics: EMNLP 2020}, pages 3441--3460.

\bibitem[{Kalamkar et~al.(2022)Kalamkar, Agarwal, Tiwari, Gupta, Karn, and Raghavan}]{kalamkar2022named}
Prathamesh Kalamkar, Astha Agarwal, Aman Tiwari, Smita Gupta, Saurabh Karn, and Vivek Raghavan. 2022.
\newblock Named entity recognition in indian court judgments.
\newblock \emph{arXiv preprint arXiv:2211.03442}.

\bibitem[{Kirchenbauer et~al.(2023)Kirchenbauer, Geiping, Wen, Katz, Miers, and Goldstein}]{kirchenbauer2023watermark}
John Kirchenbauer, Jonas Geiping, Yuxin Wen, Jonathan Katz, Ian Miers, and Tom Goldstein. 2023.
\newblock A watermark for large language models.
\newblock \emph{arXiv preprint arXiv:2301.10226}.

\bibitem[{Koreeda and Manning(2021)}]{koreeda2021contractnli}
Yuta Koreeda and Christopher~D Manning. 2021.
\newblock Contractnli: A dataset for document-level natural language inference for contracts.
\newblock \emph{arXiv preprint arXiv:2110.01799}.

\bibitem[{Leiker et~al.(2023)Leiker, Finnigan, Gyllen, and Cukurova}]{leiker2023prototyping}
Daniel Leiker, Sara Finnigan, Ashley~Ricker Gyllen, and Mutlu Cukurova. 2023.
\newblock Prototyping the use of large language models (llms) for adult learning content creation at scale.
\newblock \emph{arXiv preprint arXiv:2306.01815}.

\bibitem[{Leitner et~al.(2019)Leitner, Rehm, and Moreno-Schneider}]{leitner2019fine}
Elena Leitner, Georg Rehm, and Julian Moreno-Schneider. 2019.
\newblock Fine-grained named entity recognition in legal documents.
\newblock In \emph{International Conference on Semantic Systems}, pages 272--287. Springer.

\bibitem[{Leitner et~al.(2020)Leitner, Rehm, and Moreno-Schneider}]{leitner2020dataset}
Elena Leitner, Georg Rehm, and Juli{\'a}n Moreno-Schneider. 2020.
\newblock A dataset of german legal documents for named entity recognition.
\newblock \emph{arXiv preprint arXiv:2003.13016}.

\bibitem[{Liu et~al.(2023)Liu, Yuan, Fu, Jiang, Hayashi, and Neubig}]{liu2023pre}
Pengfei Liu, Weizhe Yuan, Jinlan Fu, Zhengbao Jiang, Hiroaki Hayashi, and Graham Neubig. 2023.
\newblock Pre-train, prompt, and predict: A systematic survey of prompting methods in natural language processing.
\newblock \emph{ACM Computing Surveys}, 55(9):1--35.

\bibitem[{Liu et~al.(2019)Liu, Ott, Goyal, Du, Joshi, Chen, Levy, Lewis, Zettlemoyer, and Stoyanov}]{liu2019roberta}
Yinhan Liu, Myle Ott, Naman Goyal, Jingfei Du, Mandar Joshi, Danqi Chen, Omer Levy, Mike Lewis, Luke Zettlemoyer, and Veselin Stoyanov. 2019.
\newblock \href {http://arxiv.org/abs/1907.11692} {Roberta: {A} robustly optimized {BERT} pretraining approach}.
\newblock \emph{CoRR}, abs/1907.11692.

\bibitem[{Luz~de Araujo et~al.(2018)Luz~de Araujo, de~Campos, de~Oliveira, Stauffer, Couto, and Bermejo}]{luz2018lener}
Pedro~Henrique Luz~de Araujo, Te{\'o}filo~E de~Campos, Renato~RR de~Oliveira, Matheus Stauffer, Samuel Couto, and Paulo Bermejo. 2018.
\newblock Lener-br: a dataset for named entity recognition in brazilian legal text.
\newblock In \emph{Computational Processing of the Portuguese Language: 13th International Conference, PROPOR 2018, Canela, Brazil, September 24--26, 2018, Proceedings 13}, pages 313--323. Springer.

\bibitem[{Mamakas et~al.(2022)Mamakas, Tsotsi, Androutsopoulos, and Chalkidis}]{mamakas2022processing}
Dimitris Mamakas, Petros Tsotsi, Ion Androutsopoulos, and Ilias Chalkidis. 2022.
\newblock \href {http://arxiv.org/abs/2211.00974} {Processing long legal documents with pre-trained transformers: Modding legalbert and longformer}.

\bibitem[{Mitchell et~al.(2023)Mitchell, Lee, Khazatsky, Manning, and Finn}]{mitchell2023detectgpt}
Eric Mitchell, Yoonho Lee, Alexander Khazatsky, Christopher~D Manning, and Chelsea Finn. 2023.
\newblock Detectgpt: Zero-shot machine-generated text detection using probability curvature.
\newblock \emph{arXiv preprint arXiv:2301.11305}.

\bibitem[{M{\o}ller et~al.(2023)M{\o}ller, Dalsgaard, Pera, and Aiello}]{moller2023prompt}
Anders~Giovanni M{\o}ller, Jacob~Aarup Dalsgaard, Arianna Pera, and Luca~Maria Aiello. 2023.
\newblock Is a prompt and a few samples all you need? using gpt-4 for data augmentation in low-resource classification tasks.
\newblock \emph{arXiv preprint arXiv:2304.13861}.

\bibitem[{Niklaus et~al.(2023)Niklaus, Matoshi, St{\"u}rmer, Chalkidis, and Ho}]{niklaus2023multilegalpile}
Joel Niklaus, Veton Matoshi, Matthias St{\"u}rmer, Ilias Chalkidis, and Daniel~E Ho. 2023.
\newblock Multilegalpile: A 689gb multilingual legal corpus.
\newblock \emph{arXiv preprint arXiv:2306.02069}.

\bibitem[{Nyffenegger et~al.(2023)Nyffenegger, Stürmer, and Niklaus}]{nyffenegger2023anonymity}
Alex Nyffenegger, Matthias Stürmer, and Joel Niklaus. 2023.
\newblock \href {http://arxiv.org/abs/2308.11103} {Anonymity at risk? assessing re-identification capabilities of large language models}.

\bibitem[{OpenAI(2023)}]{openai2023gpt4}
OpenAI. 2023.
\newblock \href {http://arxiv.org/abs/2303.08774} {Gpt-4 technical report}.

\bibitem[{Ouyang et~al.(2022)Ouyang, Wu, Jiang, Almeida, Wainwright, Mishkin, Zhang, Agarwal, Slama, Ray et~al.}]{ouyang2022training}
Long Ouyang, Jeffrey Wu, Xu~Jiang, Diogo Almeida, Carroll Wainwright, Pamela Mishkin, Chong Zhang, Sandhini Agarwal, Katarina Slama, Alex Ray, et~al. 2022.
\newblock Training language models to follow instructions with human feedback.
\newblock \emph{Advances in Neural Information Processing Systems}, 35:27730--27744.

\bibitem[{P{\u{a}}iș et~al.(2021)P{\u{a}}iș, Mitrofan, Gasan, Coneschi, and Ianov}]{puaiș2021named}
Vasile P{\u{a}}iș, Maria Mitrofan, Carol~Luca Gasan, Vlad Coneschi, and Alexandru Ianov. 2021.
\newblock Named entity recognition in the romanian legal domain.
\newblock In \emph{Proceedings of the Natural Legal Language Processing Workshop 2021}, pages 9--18.

\bibitem[{Puri et~al.(2020)Puri, Spring, Patwary, Shoeybi, and Catanzaro}]{puri2020training}
Raul Puri, Ryan Spring, Mostofa Patwary, Mohammad Shoeybi, and Bryan Catanzaro. 2020.
\newblock Training question answering models from synthetic data.
\newblock \emph{arXiv preprint arXiv:2002.09599}.

\bibitem[{Radford et~al.(2019)Radford, Wu, Child, Luan, Amodei, Sutskever et~al.}]{radford2019language}
Alec Radford, Jeffrey Wu, Rewon Child, David Luan, Dario Amodei, Ilya Sutskever, et~al. 2019.
\newblock Language models are unsupervised multitask learners.
\newblock \emph{OpenAI blog}, 1(8):9.

\bibitem[{Rosenbaum et~al.(2022{\natexlab{a}})Rosenbaum, Soltan, Hamza, Saffari, Damonte, and Groves}]{rosenbaum2022clasp}
Andy Rosenbaum, Saleh Soltan, Wael Hamza, Amir Saffari, Macro Damonte, and Isabel Groves. 2022{\natexlab{a}}.
\newblock Clasp: Few-shot cross-lingual data augmentation for semantic parsing.
\newblock \emph{arXiv preprint arXiv:2210.07074}.

\bibitem[{Rosenbaum et~al.(2022{\natexlab{b}})Rosenbaum, Soltan, Hamza, Versley, and Boese}]{rosenbaum2022linguist}
Andy Rosenbaum, Saleh Soltan, Wael Hamza, Yannick Versley, and Markus Boese. 2022{\natexlab{b}}.
\newblock Linguist: Language model instruction tuning to generate annotated utterances for intent classification and slot tagging.
\newblock \emph{arXiv preprint arXiv:2209.09900}.

\bibitem[{Sanh et~al.(2019)Sanh, Debut, Chaumond, and Wolf}]{sanh2019distilbert}
Victor Sanh, Lysandre Debut, Julien Chaumond, and Thomas Wolf. 2019.
\newblock \href {http://arxiv.org/abs/1910.01108} {Distilbert, a distilled version of {BERT:} smaller, faster, cheaper and lighter}.
\newblock \emph{CoRR}, abs/1910.01108.

\bibitem[{Savelka et~al.(2023{\natexlab{a}})Savelka, Ashley, Gray, Westermann, and Xu}]{savelka2023can}
Jaromir Savelka, Kevin~D Ashley, Morgan~A Gray, Hannes Westermann, and Huihui Xu. 2023{\natexlab{a}}.
\newblock Can gpt-4 support analysis of textual data in tasks requiring highly specialized domain expertise?
\newblock \emph{arXiv preprint arXiv:2306.13906}.

\bibitem[{Savelka et~al.(2023{\natexlab{b}})Savelka, Ashley, Gray, Westermann, and Xu}]{savelka2023explaining}
Jaromir Savelka, Kevin~D Ashley, Morgan~A Gray, Hannes Westermann, and Huihui Xu. 2023{\natexlab{b}}.
\newblock Explaining legal concepts with augmented large language models (gpt-4).
\newblock \emph{arXiv preprint arXiv:2306.09525}.

\bibitem[{Semo et~al.(2022)Semo, Bernsohn, Hagag, Hayat, and Niklaus}]{semo2022classactionprediction}
Gil Semo, Dor Bernsohn, Ben Hagag, Gila Hayat, and Joel Niklaus. 2022.
\newblock Classactionprediction: A challenging benchmark for legal judgment prediction of class action cases in the us.
\newblock \emph{arXiv preprint arXiv:2211.00582}.

\bibitem[{Silva et~al.(2020)Silva, Gonçalves, Godinho, Antunes, and Curado}]{9162683}
Paulo Silva, Carolina Gonçalves, Carolina Godinho, Nuno Antunes, and Marilia Curado. 2020.
\newblock \href {https://doi.org/10.1109/INFOCOMWKSHPS50562.2020.9162683} {Using nlp and machine learning to detect data privacy violations}.
\newblock In \emph{IEEE INFOCOM 2020 - IEEE Conference on Computer Communications Workshops (INFOCOM WKSHPS)}, pages 972--977.

\bibitem[{Skylaki et~al.(2020)Skylaki, Oskooei, Bari, Herger, and Kriegman}]{skylaki2020named}
Stavroula Skylaki, Ali Oskooei, Omar Bari, Nadja Herger, and Zac Kriegman. 2020.
\newblock Named entity recognition in the legal domain using a pointer generator network.
\newblock \emph{arXiv preprint arXiv:2012.09936}.

\bibitem[{Sun et~al.(2020)Sun, Qiu, Xu, and Huang}]{sun2020finetune}
Chi Sun, Xipeng Qiu, Yige Xu, and Xuanjing Huang. 2020.
\newblock \href {http://arxiv.org/abs/1905.05583} {How to fine-tune bert for text classification?}

\bibitem[{Thorne et~al.(2018)Thorne, Vlachos, Christodoulopoulos, and Mittal}]{thorne2018fever}
James Thorne, Andreas Vlachos, Christos Christodoulopoulos, and Arpit Mittal. 2018.
\newblock Fever: a large-scale dataset for fact extraction and verification.
\newblock In \emph{Proceedings of the 2018 Conference of the North American Chapter of the Association for Computational Linguistics: Human Language Technologies, Volume 1 (Long Papers)}, pages 809--819.

\bibitem[{Touvron et~al.(2023)Touvron, Martin, Stone, Albert, Almahairi, Babaei, Bashlykov, Batra, Bhargava, Bhosale, Bikel, Blecher, Ferrer, Chen, Cucurull, Esiobu, Fernandes, Fu, Fu, Fuller, Gao, Goswami, Goyal, Hartshorn, Hosseini, Hou, Inan, Kardas, Kerkez, Khabsa, Kloumann, Korenev, Koura, Lachaux, Lavril, Lee, Liskovich, Lu, Mao, Martinet, Mihaylov, Mishra, Molybog, Nie, Poulton, Reizenstein, Rungta, Saladi, Schelten, Silva, Smith, Subramanian, Tan, Tang, Taylor, Williams, Kuan, Xu, Yan, Zarov, Zhang, Fan, Kambadur, Narang, Rodriguez, Stojnic, Edunov, and Scialom}]{touvron2023llama}
Hugo Touvron, Louis Martin, Kevin Stone, Peter Albert, Amjad Almahairi, Yasmine Babaei, Nikolay Bashlykov, Soumya Batra, Prajjwal Bhargava, Shruti Bhosale, Dan Bikel, Lukas Blecher, Cristian~Canton Ferrer, Moya Chen, Guillem Cucurull, David Esiobu, Jude Fernandes, Jeremy Fu, Wenyin Fu, Brian Fuller, Cynthia Gao, Vedanuj Goswami, Naman Goyal, Anthony Hartshorn, Saghar Hosseini, Rui Hou, Hakan Inan, Marcin Kardas, Viktor Kerkez, Madian Khabsa, Isabel Kloumann, Artem Korenev, Punit~Singh Koura, Marie-Anne Lachaux, Thibaut Lavril, Jenya Lee, Diana Liskovich, Yinghai Lu, Yuning Mao, Xavier Martinet, Todor Mihaylov, Pushkar Mishra, Igor Molybog, Yixin Nie, Andrew Poulton, Jeremy Reizenstein, Rashi Rungta, Kalyan Saladi, Alan Schelten, Ruan Silva, Eric~Michael Smith, Ranjan Subramanian, Xiaoqing~Ellen Tan, Binh Tang, Ross Taylor, Adina Williams, Jian~Xiang Kuan, Puxin Xu, Zheng Yan, Iliyan Zarov, Yuchen Zhang, Angela Fan, Melanie Kambadur, Sharan Narang, Aurelien Rodriguez, Robert Stojnic, Sergey Edunov, and Thomas
  Scialom. 2023.
\newblock \href {http://arxiv.org/abs/2307.09288} {Llama 2: Open foundation and fine-tuned chat models}.

\bibitem[{Tsarapatsanis and Aletras(2021)}]{tsarapatsanis2021ethical}
Dimitrios Tsarapatsanis and Nikolaos Aletras. 2021.
\newblock On the ethical limits of natural language processing on legal text.
\newblock In \emph{Findings of the Association for Computational Linguistics: ACL-IJCNLP 2021}, pages 3590--3599.

\bibitem[{Vaswani et~al.(2017)Vaswani, Shazeer, Parmar, Uszkoreit, Jones, Gomez, Kaiser, and Polosukhin}]{vaswani2017}
Ashish Vaswani, Noam Shazeer, Niki Parmar, Jakob Uszkoreit, Llion Jones, Aidan~N. Gomez, Lukasz Kaiser, and Illia Polosukhin. 2017.
\newblock \href {http://arxiv.org/abs/1706.03762} {Attention is all you need}.
\newblock \emph{CoRR}, abs/1706.03762.

\bibitem[{Veselovsky et~al.(2023)Veselovsky, Ribeiro, Arora, Josifoski, Anderson, and West}]{veselovsky2023generating}
Veniamin Veselovsky, Manoel~Horta Ribeiro, Akhil Arora, Martin Josifoski, Ashton Anderson, and Robert West. 2023.
\newblock Generating faithful synthetic data with large language models: A case study in computational social science.
\newblock \emph{arXiv preprint arXiv:2305.15041}.

\bibitem[{Veyseh et~al.(2021)Veyseh, Lai, Dernoncourt, and Nguyen}]{veyseh2021unleash}
Amir Pouran~Ben Veyseh, Viet Lai, Franck Dernoncourt, and Thien~Huu Nguyen. 2021.
\newblock Unleash gpt-2 power for event detection.
\newblock In \emph{Proceedings of the 59th Annual Meeting of the Association for Computational Linguistics and the 11th International Joint Conference on Natural Language Processing (Volume 1: Long Papers)}, pages 6271--6282.

\bibitem[{Wang et~al.(2019)Wang, Yang, and Xia}]{wang2019simplified}
Juan Wang, Yongyi Yang, and Bin Xia. 2019.
\newblock A simplified cohen’s kappa for use in binary classification data annotation tasks.
\newblock \emph{IEEE Access}, 7:164386--164397.

\bibitem[{Wei et~al.(2021)Wei, Bosma, Zhao, Guu, Yu, Lester, Du, Dai, and Le}]{wei2021finetuned}
Jason Wei, Maarten Bosma, Vincent~Y Zhao, Kelvin Guu, Adams~Wei Yu, Brian Lester, Nan Du, Andrew~M Dai, and Quoc~V Le. 2021.
\newblock Finetuned language models are zero-shot learners.
\newblock \emph{arXiv preprint arXiv:2109.01652}.

\bibitem[{Yu et~al.(2020)Yu, Guo, Zhang, Li, Ji, Tang, and Wu}]{9206907}
Yaoquan Yu, Yuefeng Guo, Zhiyuan Zhang, Mengshi Li, Tianyao Ji, Wenhu Tang, and Qinghua Wu. 2020.
\newblock \href {https://doi.org/10.1109/IJCNN48605.2020.9206907} {Intelligent classification and automatic annotation of violations based on neural network language model}.
\newblock In \emph{2020 International Joint Conference on Neural Networks (IJCNN)}, pages 1--7.

\end{thebibliography}

\appendix

\section{Experiments Setting}

All experiments were conducted on AWS g5.4xlarge instance, equipped with $1$ NVIDIA A10G GPU. The total GPU hours are 85. For each model, the reported metrics are obtained by computing the mean and standard deviation across five runs with randomly initialized weights. All code\footnote{\url{https://github.com/darrow-labs/LegalLens}} and  datasets (NER\footnote{\url{https://huggingface.co/datasets/darrow-ai/LegalLensNER}} and NLI\footnote{\url{https://huggingface.co/datasets/darrow-ai/LegalLensNLI}}) are available.

\subsection{Library Versions}
We used the following libraries and associated
versions: python 3.8, transformers 4.31.0, seqeval 1.2.2, streamlit 1.25.0, datasets 2.14.2, evaluate 0.4.0, wandb 0.15.7, torch 2.0.1, accelerate 0.21.0, sentencepiece 0.1.99, google cloud aiplatform 1.28.1, openai 0.27.8, langchain 0.0.248, ipython 8.12.2, typer 0.9.0, nltk 3.8, matplotlib 3.7.2.

\begin{figure*}
    \centering
    \includegraphics[width=1.4\linewidth]{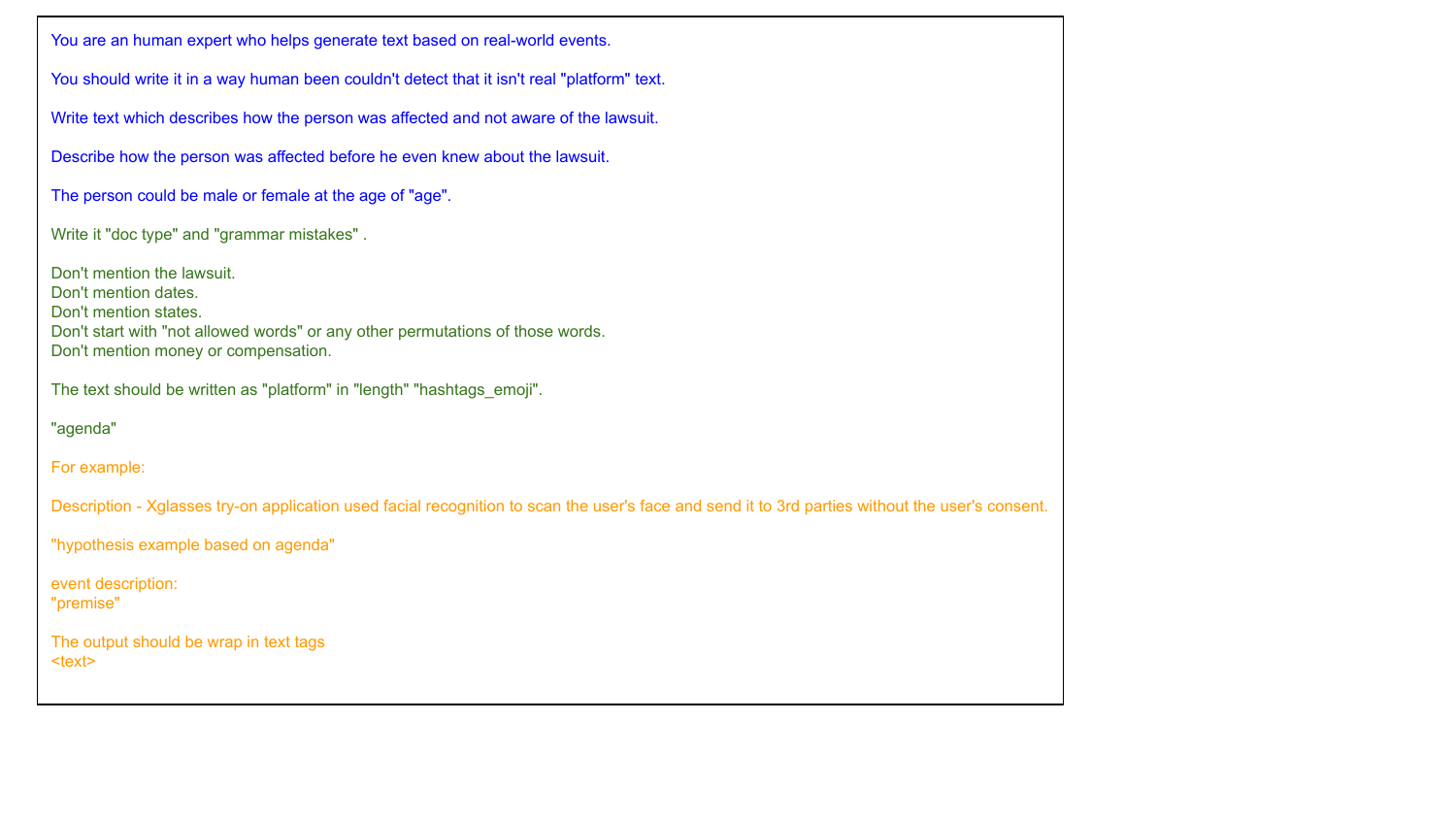}
    \caption{Prompt design for generating NLI data set. Prompt contains the \textcolor{blue}{task description}, \textcolor{green}{specific instructions}, and \textcolor{orange}{few-shot examples}.}
\label{figure:NLI_datagen_prompts}
\end{figure*}


\section{Annotation Platform}
\label{sec:annotation}

\begin{figure*}[!ht]
\includegraphics[width=\textwidth]{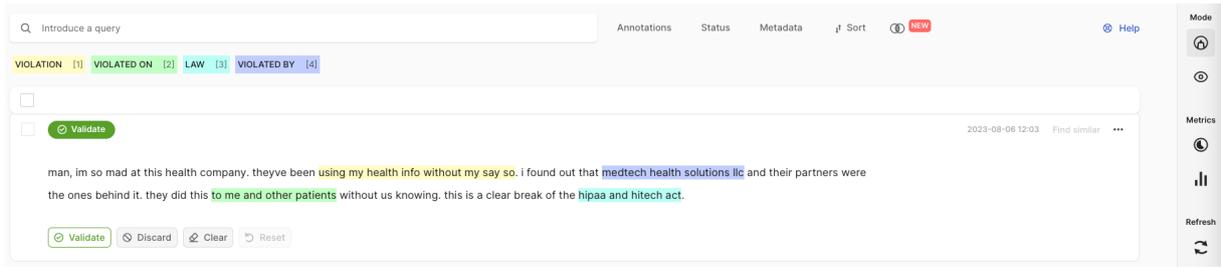}
\caption{The platform for the human annotations.}
\label{fig:annotation_plattform}
\end{figure*}

We ran our annotation platform with the Argilla library \footnote{\url{https://github.com/argilla-io/argilla}} available under an Apache-2.0 license.

Figure \ref{fig:annotation_plattform} shows a screenshot of the annotation platform our human experts used.

\section{Data Distribution}
\begin{figure*}[!ht]
\includegraphics[width=\textwidth]{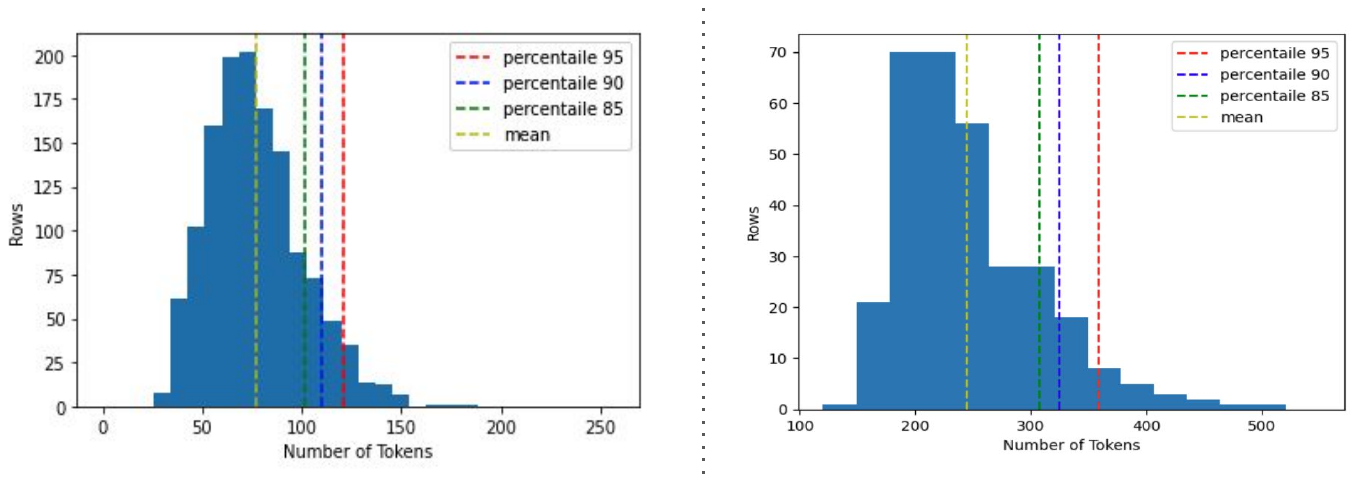}
\caption{Token Distributions for (left) NER and (right) NLI.}
\label{fig:Token_Distribution}
\end{figure*}

Figure \ref{fig:Token_Distribution} shows the datasets tokens distribution.
\label{sec:data_destribution}

\begin{table}[h]
\centering
\resizebox{0.5\textwidth}{!}{%
\begin{tabular}{|c|p{3.5cm}|c|}
\hline
\textbf{Entity} & \textbf{Description} & \textbf{\# Labeled Samples} \\
\hline
LAW & Specific law or regulation breached. & 292 \\
\hline
VIOLATION & Content describing the violation. & 1326 \\
\hline
VIOLATED BY & Entity committing the violation. & 292 \\
\hline
VIOLATED ON & Victim or affected party. & 292 \\
\hline
\end{tabular}
}
\caption{Distribution of the NER entities produced by the generation process (2202 in total).}
\label{table:ner_entities}
\end{table}

\begin{table}[h]
\centering
\resizebox{0.5
\textwidth}{!}{%
\begin{tabular}{|c|p{3.5cm}|c|c|}
\hline
\textbf{Entity} & 
\textbf{Description} & 
\textbf{Labels} & 
\textbf{\# Labeled Samples} \\
\hline
Consumer Protection & Deceptive advertising, fraud and unfair business practices. &16/17/29& 62 \\
\hline
Privacy & Unauthorized collection, use, or disclosure of personal data.&56/54/53 & 163 \\
\hline
TCPA & Unauthorized telemarketing calls, faxes and text messages.&26/27/21 & 74 \\
\hline
Wage & Illegal underpayment and unfair compensation practices by employers.&6/3/4 & 13 \\
\hline
\end{tabular}
}
\caption{Distribution of labeled samples across various legal domains for the NLI task. The number of samples is in the format of Contradiction/Neutral/Entailment.}
\label{table:NLI_DOMAINS}
\end{table}

\section{Prompts}

\begin{figure*}[ht]
    \centering
    \begin{subfigure}{\textwidth}
        \centering
        \includegraphics[width=0.9\linewidth]{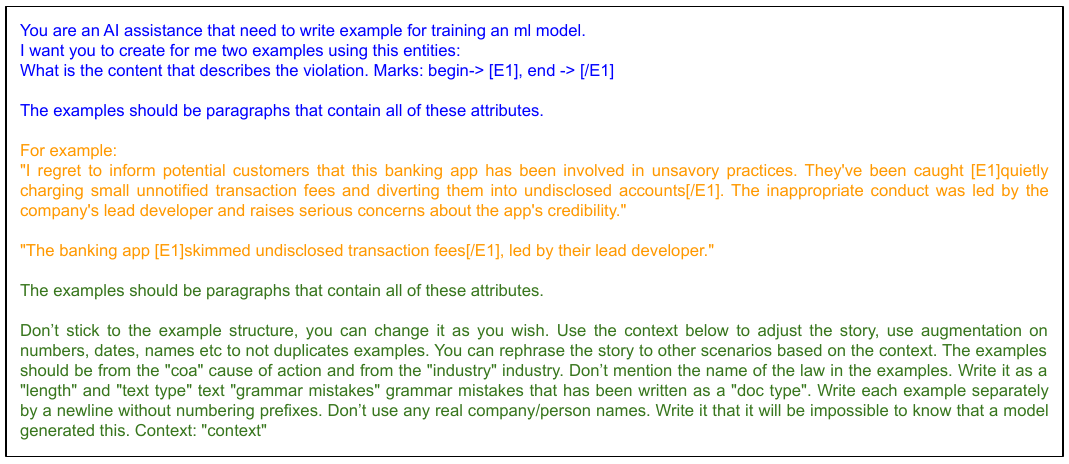}
        \caption{Prompt design for Implicit NER data set. Prompt contains the \textcolor{blue}{task description}, \textcolor{orange}{few-shot examples}, and \textcolor{green}{specific instructions}.}
    \end{subfigure}
    \begin{subfigure}{\textwidth}
    \includegraphics[width=1.35\linewidth]{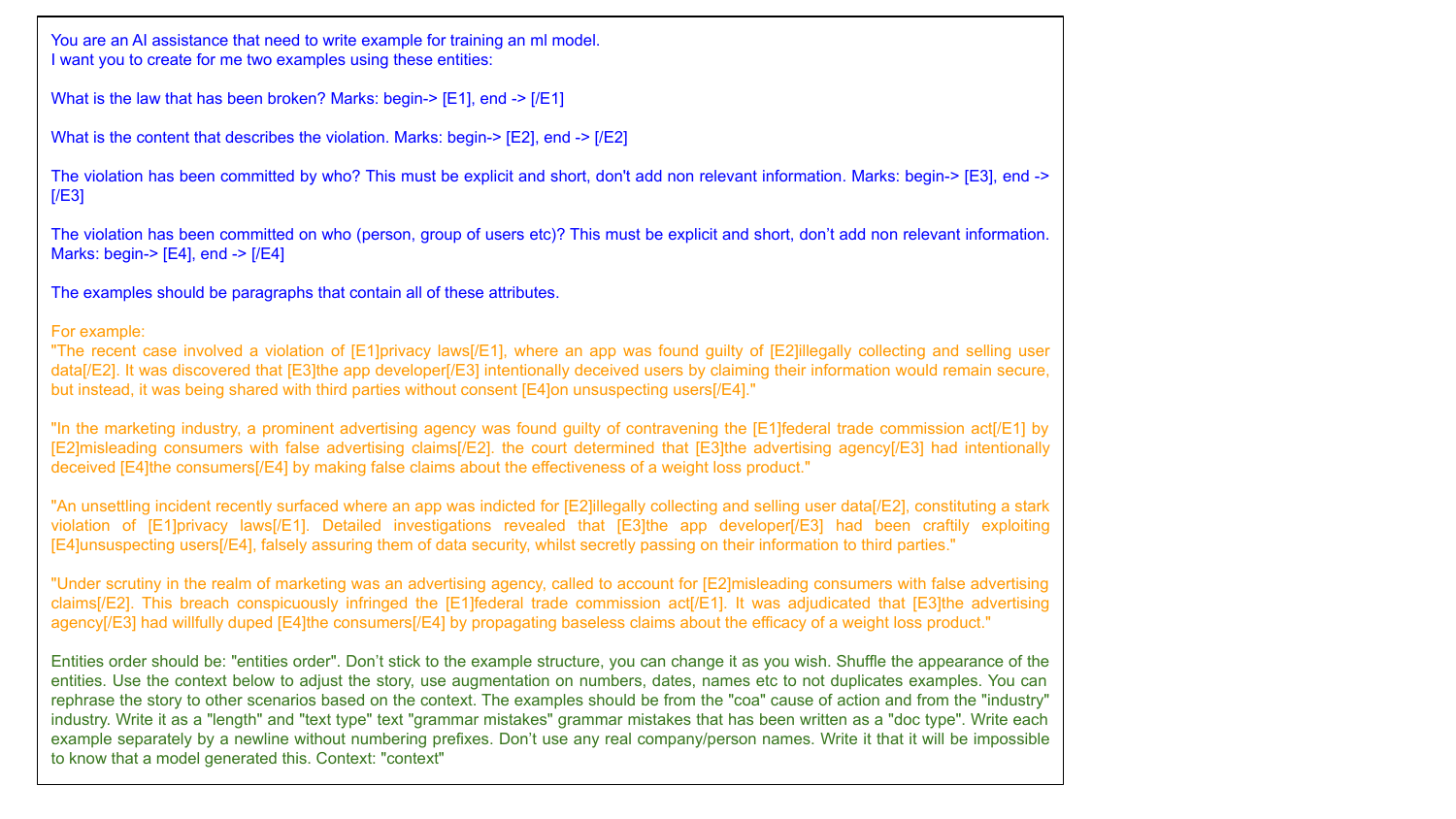}
    \caption{Prompt design for Explicit NER data set. Prompt contains the \textcolor{blue}{task description}, \textcolor{orange}{few-shot examples}, and \textcolor{green}{specific instructions}.}
    \end{subfigure}
    \caption{The prompts used for generating the NER data set.}
\label{figure:NER_datagen_prompts}
\end{figure*}

\begin{figure*}[ht]
    \centering
    \begin{subfigure}{\textwidth}
        \centering
        \includegraphics[width=\linewidth]{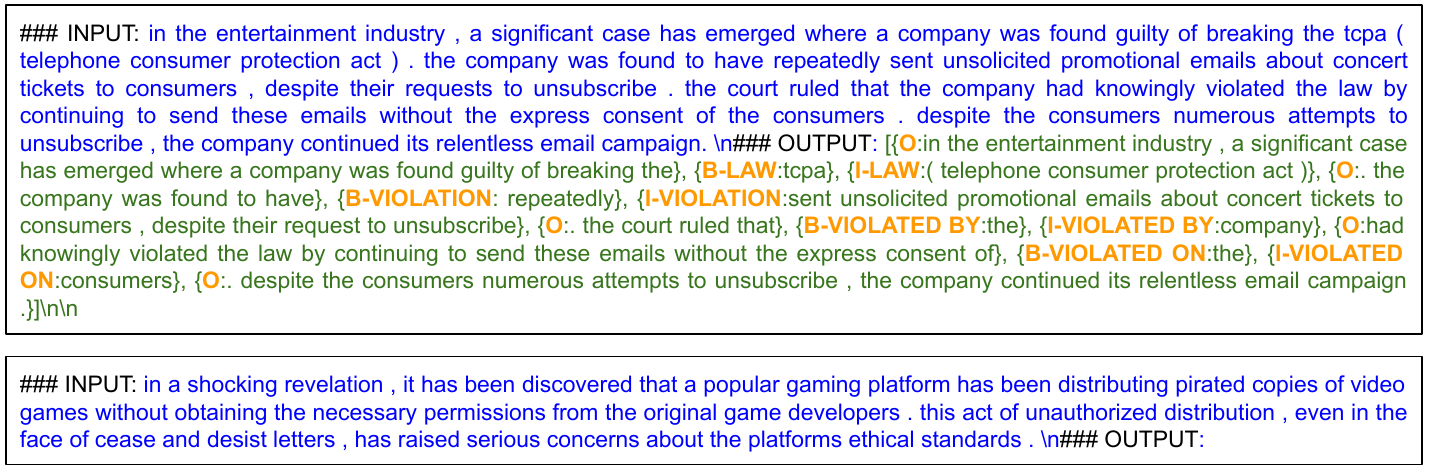}
        \caption{Prompt design for NER. (Top) Training prompt, containing the input and output tags, \textcolor{blue}{input text}, \textcolor{green}{output text} and corresponding \textcolor{orange}{NER tags}. (Bottom) Inference prompt, containing only the input and output tags, \textcolor{blue}{input text}.}
    \end{subfigure}
    \begin{subfigure}{\textwidth}
    \centering
    \includegraphics[width=\linewidth]{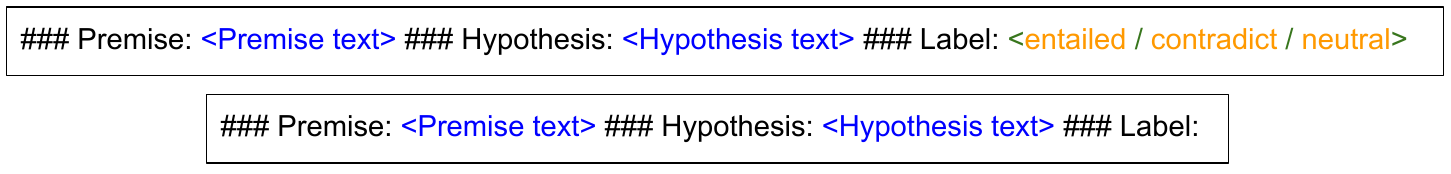}
    \caption{Prompt design for NLI. (Top) Training prompt, containing the input and output tags, \textcolor{blue}{premise} and \textcolor{blue}{hypothesis} texts, and corresponding \textcolor{orange}{labels}. (Bottom) Inference prompt, containing relevant tags, and \textcolor{blue}{premise} and \textcolor{blue}{hypothesis} texts.}
    \end{subfigure}
    \caption{The prompts used for fine-tuning open-source LLMs across (a) NER and (b) NLI tasks.}
\label{figure:llm-prompts}
\end{figure*}

\begin{figure*}
\centering
\includegraphics[width=\linewidth]{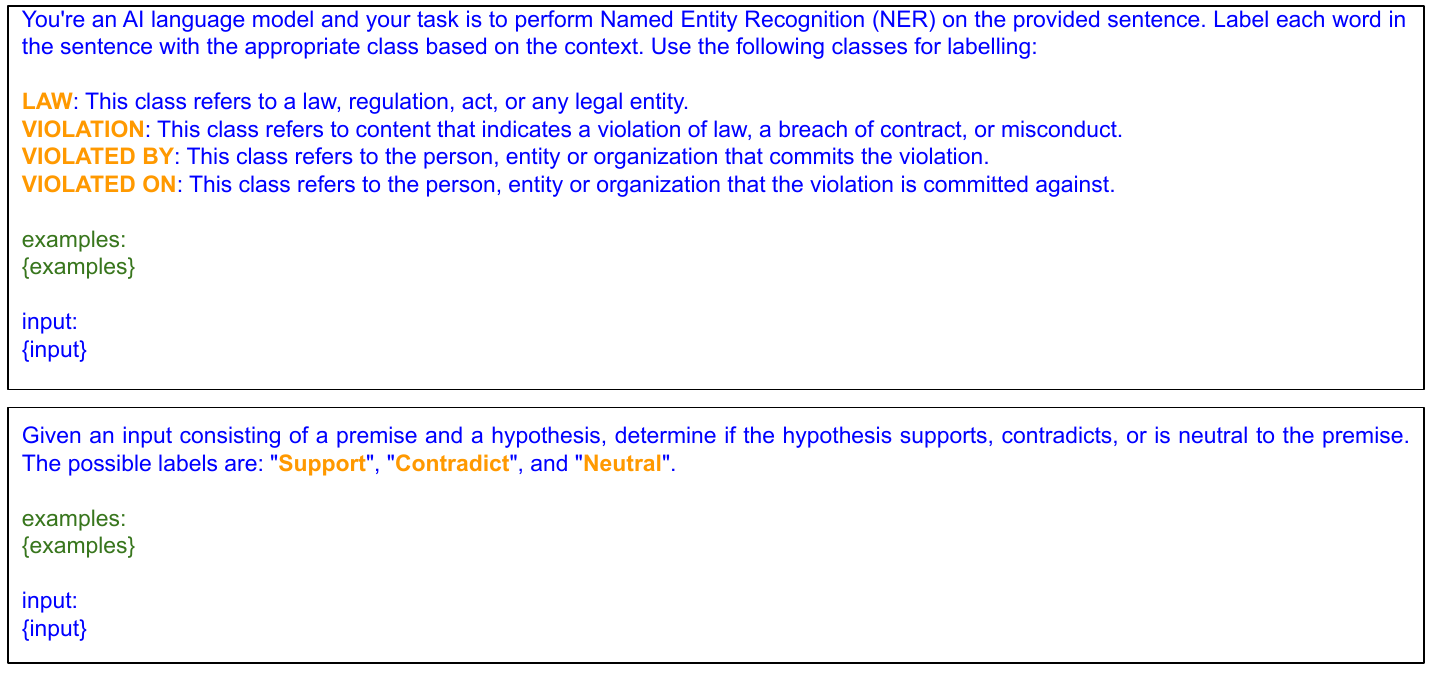}
\caption{Few-shot prompt designs for (top) NER and (below) NLI experiments using OpenAI GPT models. Prompts contain \textcolor{blue}{input}, \textcolor{blue}{general task-specific instructions}, \textcolor{orange}{labels} for each task and \textcolor{green}{few-shot examples}.}
\label{figure:few-shot_prompts}
\end{figure*}

In this appendix, we detail the data generation prompts utilized for the GPT-4 model. The prompts for the datasets creation are illustrated in Figures \ref{figure:NER_datagen_prompts} and \ref{figure:NLI_datagen_prompts}. Meanwhile, the prompts for fine-tuning can be found in Figure \ref{figure:llm-prompts}. The prompt for the Few-shot approach is depicted in Figure \ref{figure:few-shot_prompts}


\begin{figure*}
    \centering
    \includegraphics[width=1\linewidth]{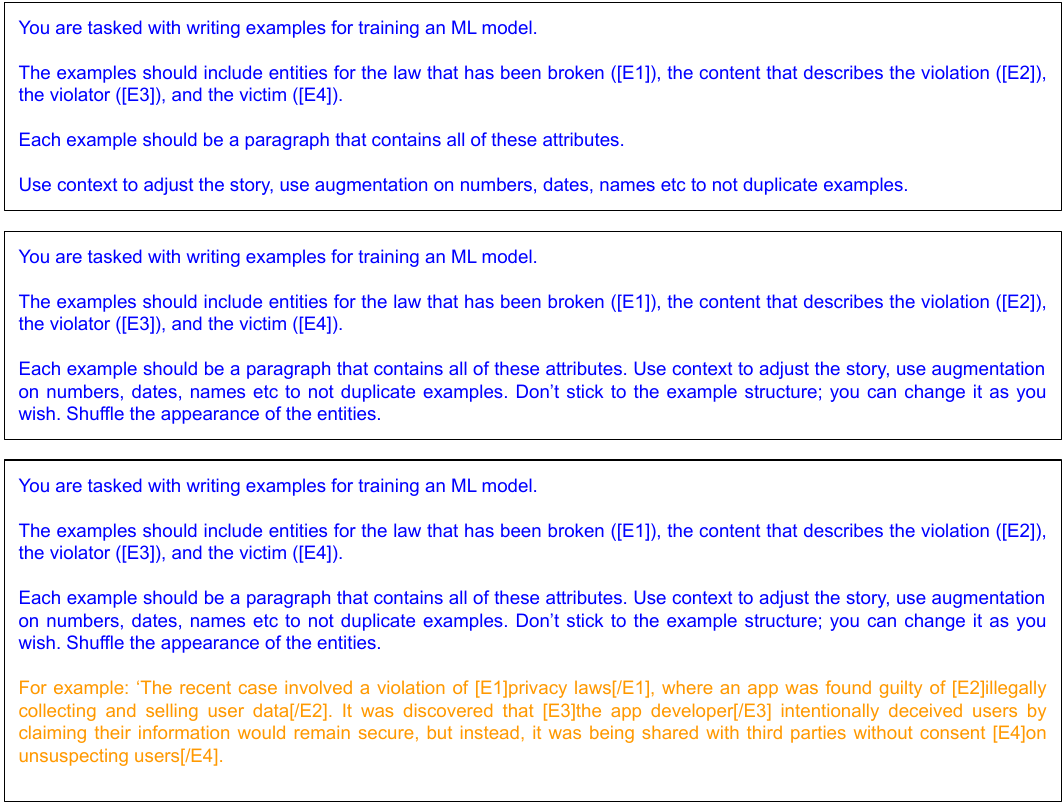}
    \caption{(Top to Bottom) Iterations of prompt design for generating the Explicit NER data set. Prompts contain the \textcolor{blue}{task description}, and \textcolor{orange}{few-shot examples}. Figure~\ref{figure:NER_datagen_prompts}-b contains the final version of prompt used.}
\label{figure:prompt_iterations}
\end{figure*}

\end{document}